\documentclass[pdflatex]{nolta}
\usepackage{mathtools}
\usepackage{bm}

\Vol{17}%
\No{1}%
\Year{2026}%
\Month{1}%
\PaperID{2026ENP0886}

\title{Quantization robustness from dense representations of
  sparse functions in high-capacity kernel associative memory}

\AUTHOR{
\author{Akira Tamamori}{1}\orcid{0009-0000-8893-0058}
}

\AFFILIATE{ \affiliate{Faculty of Information Science, Aichi
    Institute of Technology\\1247 Yachigusa, Yakusa-cho, Toyota-shi, Aichi 470-0392, Japan}{1} }

\received{10}{27}{20XX}
\revised{12}{29}{20XX}
\published{7}{1}{20XX}

\begin{document}

\begin{abstract}
  High-capacity associative memories based on Kernel Logistic
  Regression (KLR) achieve strong retrieval performance but typically
  require substantial computational resources. This paper investigates
  the compressibility of KLR Hopfield networks to clarify the
  geometric principles underlying their robust representations. We
  present a geometric interpretation based on spontaneous symmetry
  breaking and Walsh analysis, and examine it through compression
  experiments involving quantization and pruning. The experiments
  reveal a clear asymmetry: the network remains robust under
  low-precision quantization while exhibiting strong sensitivity to
  pruning. We interpret this behavior through a ``sparse function,
  dense representation'' principle, in which a sparse input mapping is
  implemented through a dense bimodal parameterization. These findings
  suggest a practical route toward hardware-efficient kernel
  associative memories and provide insight into the geometric
  principles underlying robust representation in neural systems.
\end{abstract}

\maketitle

\section{Introduction}

Associative memory models, epitomized by the Hopfield network, have
recently attracted renewed interest through the integration of modern
machine learning techniques. In particular, our recent work showed
that Kernel Logistic Regression (KLR) learning substantially enhances
storage capacity and robustness, achieving performance beyond
classical limits~\cite{tamamori2025, tamamori2025b}. These
``kernelized'' memories operate by embedding patterns into a
high-dimensional feature space, forming deep attractor
basins. Furthermore, we theoretically identified a geometric
self-organization mechanism, termed \textit{Spectral
  Concentration}~\cite{tamamori2025c}, that underlies this
high-capacity behavior.

In parallel with these developments, model compression has become a
major research topic in deep learning. Recent advances in Large
Language Models have demonstrated that aggressive quantization, such
as 4-bit or even 1.58-bit representations (e.g., BitNet~\cite{Ma2024}),
can preserve performance while substantially reducing computational
cost. Whether similar compression principles apply to kernel-based
associative memories, however, remains largely unexplored.

Addressing this question is important because the high performance of
KLR Hopfield networks comes at a substantial computational
cost. Retrieval in kernel networks requires computing weighted sums
over all stored patterns using high-precision dual variables, leading
to considerable computational and memory overhead. This poses a
practical challenge for deployment on resource-constrained edge
devices and neuromorphic hardware, where memory bandwidth and energy
efficiency are critical. In contrast, biological neural systems are
known to operate efficiently with noisy, low-precision synaptic
weights and distributed representations. This raises a central
question: Does high-capacity associative memory fundamentally require
high-precision continuous parameters, or can the essential information
be represented in a more robust and discrete form? More broadly, what
geometric structure governs this representation?

In this study, we investigate the compressibility of KLR Hopfield
networks from this perspective. We systematically analyze the effects
of two forms of compression: \textit{Uniform Quantization} (reducing
bit precision) and \textit{Magnitude Pruning} (inducing sparsity). Our
experiments reveal a clear dichotomy in the network's
compressibility. The network remains highly robust to quantization,
maintaining near-perfect recall accuracy even at 2-bit resolution. In
contrast, it is highly sensitive to sparsification, with performance
degrading rapidly even under moderate pruning.

Our work provides a geometric interpretation of this dichotomy,
connecting the empirical compression behavior with the structure of
the learned representation. It should be noted that throughout this
paper, the terms ``geometry'' and ``geometric interpretation'' do not
refer to the spatial or topological arrangement of the network's
physical connections. Rather, they pertain to the geometric structure
of the parameter space (e.g., the bimodal distribution of weights) and
the intrinsic information geometry of the statistical manifold, such
as the curvature defined by the Fisher Information
Matrix~\cite{Amari2016}. The main contributions of this paper are as
follows:

\begin{enumerate}
\item We identify a dichotomy in the compressibility of KLR Hopfield
  networks: robustness to quantization and sensitivity to sparsity.

\item We develop a unified geometric interpretation of this behavior
  based on two complementary observations: the emergence of a bimodal,
  digital-like weight distribution and a Walsh analysis that reveals
  sparse functional structure together with distributed
  parameterization.

\item We support this interpretation through extensive experiments and
  discuss its implications for hardware-efficient neuromorphic
  computing.
\end{enumerate}

The remainder of this paper is organized as follows.
Section~\ref{sec:related_work} reviews related work on model
compression and distributed representations.
Section~\ref{sec:phenomena} introduces the experimental setting and
presents the empirical contrast between quantization robustness and
sparsity sensitivity.  Section~\ref{sec:theory} develops the geometric
interpretation of the learned representation, including the origin of
the bimodal distribution, the Walsh influence analysis, and the
scaling behavior under quantization.  Section~\ref{sec:validation}
provides additional experimental validation of the proposed
interpretation.  Finally, Section~\ref{sec:discussion} discusses
broader implications of the results, and Section~\ref{sec:conclusion}
concludes the paper.

\section{Related Work}
\label{sec:related_work}
Our study on compressing kernel associative memories relates to three
research areas: low-precision deep learning, distributed
representations, and the neurobiology of synaptic precision.

\subsection{Low-Precision Deep Learning}
Low-precision inference has been widely studied in deep learning.
Binarized Neural Networks have shown that weights can be reduced to
1-bit without substantial loss of accuracy in complex
tasks~\cite{Hubara2016, Rastegari2016}. More recently, 1.58-bit
quantization in Large Language Models (e.g., BitNet~\cite{Ma2024}) has
demonstrated the effectiveness of extreme compression. Our work
extends these results from feedforward models to recurrent,
kernel-based associative memories, showing that comparable robustness
can be achieved.

\subsection{Holographic and Distributed Representations}
The concept of ``holographic'' representation, in which information is
distributed rather than localized, has a long history. It is related
to Holographic Reduced Representations~\cite{Plate1995} and the
broader framework of Vector Symbolic Architectures~\cite{Gayler2003,
  Kanerva2009}. In these models, information is encoded across
high-dimensional vectors, enabling superposition and retrieval.

In this paper, we define a representation as ``holographic'' if it
satisfies two properties: (1) information is distributed across
parameters, making the system sensitive to sparsity, and (2) the
system is robust to local perturbations, such as low-precision
quantization.  Our results suggest that KLR learning yields a coding
scheme consistent with these properties.

\subsection{Biological Plausibility of Low-Precision Synapses}
From a physiological perspective, the present results are consistent
with principles of neural computation in the brain. Biological systems
operate with noisy and imprecise components. Bartol et
al.~\cite{Bartol2015} provided nanoconnectomic evidence that
hippocampal synapses operate with low precision (approximately 4.7
bits). This observation suggests that biological systems, similar to
the KLR model considered here, rely on low-precision distributed
representations rather than high-precision localized encoding.

\section{Preliminaries and Empirical Phenomena}
\label{sec:phenomena}
In this section, we briefly review the kernel Hopfield network model
and the compression techniques used in this study. We then present the
empirical observations that motivate the theoretical analysis,
focusing on the contrast between robustness to quantization and
sensitivity to sparsity.

\subsection{Kernel Hopfield Network and the Ridge of Optimization}
The network consists of $N$ neurons with state
$\bm{s} \in \{-1, 1\}^{N}$, storing $P$ patterns
$\{\boldsymbol{\xi}^{\mu}\}_{\mu=1}^{P}$. In our primary experiments,
each element of the patterns is generated independently from a
symmetric Bernoulli distribution, taking values in $\{-1, 1\}$ with
equal probability ($p=0.5$). This ensures that the input data
distribution is entirely dense and uncorrelated.  The dual variables
$(\alpha_{\mu i})_{1\leq\mu\leq P, \: 1\leq i\leq N}$ are learned via
Kernel Logistic Regression (KLR)~\cite{Schölkopf2001} using the RBF
kernel $K(\bm{x}, \bm{y}) = \exp(-\gamma\|\bm{x} - \bm{y}\|^{2})$. The
retrieval dynamics are governed by the input potential:
\begin{equation}
h_i(\bm{s}) = \sum_{\mu=1}^P \alpha_{\mu i} K(\bm{s}, \boldsymbol{\xi}^\mu).
\end{equation}

For the experiments, we focus on a regime identified in previous work,
termed the ``Ridge of Optimization''~\cite{tamamori2025c}. This region
corresponds to a narrow band in the $(\gamma,P/N)$ phase space where
attractor stability is maximized under high-load conditions. Unless
otherwise noted, we use $P/N=3.0$ and $\gamma=0.02$, which were shown
to yield both high capacity and stability.
 
\subsection{Compression Schemes and Performance Metrics}

\subsubsection{Compression Schemes}
To evaluate redundancy in the learned weights $\alpha_{\mu i}$, we
apply two compression techniques post-training.

\textbf{1. Uniform Quantization:} We apply element-wise uniform
quantization to the weight matrix $\bm{A}\coloneqq (\alpha_{\mu i})$.
Let $x$ denote a weight element. The quantized value $Q_k(x)$ with
$k$-bit precision is defined as:
\begin{equation}
  Q_k(x) = \Delta \cdot \mathrm{round}\left(
    \frac{x - x_{\min}}{\Delta}
  \right) + x_{\min},
\end{equation}
where $x_{\min}$ and $x_{\max}$ denote the minimum and maximum values
of $\bm{A}$, and $\Delta = (x_{\max} - x_{\min}) / (2^k - 1)$. We
evaluate $k\in \{32, 16, \ldots, 2, 1\}$. For $k=1$, we use the sign of
the weights relative to the mean or median.

\textbf{2. Magnitude Pruning (Sparsification):} We induce sparsity by
removing weights with small magnitude. It is important to clarify that
throughout this paper, the term ``sparsity'' refers exclusively to the
properties of the learned representation (i.e., the weight matrix
$\bm{A}$ or the functional input-output mapping), and not to the
sparsity of the input patterns themselves.  For a target sparsity
ratio $S \in [0, 1)$, we define a threshold $\tau$ such that a
fraction $S$ of the elements in $\bm{A}$ satisfy
$|\alpha_{\mu i}| < \tau$. The pruned matrix $\bm{A}_{p}$ is defined
as:
\begin{equation}
  \bm{A}_{p} = \bm{A} \odot \mathbb{I}(|\bm{A}| \ge \tau),
\end{equation}
where $|\cdot|$ and $\ge$ are applied element-wise, $\odot$ denotes
the Hadamard product, and $\mathbb{I}(\cdot)$ is the indicator
function.

\subsubsection{Performance Metrics}
\begin{itemize}
\item \textbf{Bit Accuracy:} The proportion of neurons whose updated
  states correctly match the corresponding stored pattern
  $\boldsymbol{\xi}^{\mu}$ after a single update step from
  $\bm{s} = \boldsymbol{\xi}^{\mu}$.

\item \textbf{Bit-wise Recall Accuracy:} The proportion of neurons in
  the converged state that correctly match the target pattern when the
  dynamics are initialized from a noisy input.

\item \textbf{Stability Margin:} The mean alignment
  $h_i(\boldsymbol{\xi}^{\mu}) \xi_i^{\mu}$, where larger positive
  values indicate greater robustness to perturbations.
\end{itemize}

\subsection{Experimental Setup}
All simulations were implemented in Python 3.13 using the
\verb+NumPy+ 2.1.3 and \verb+SciPy+ 1.15.2 libraries. The experiments
were conducted on a workstation equipped with an Intel Core
i9-9900K CPU and 64\,GB of RAM. No GPU acceleration was used.

For all primary experiments, the number of neurons was fixed at
$N=100$. The number of stored patterns $P$ was determined by the
specified load ratio; for example, at $P/N=3.0$, the network stores
$P=300$ patterns. Unless otherwise stated, all reported results are
averaged over 10 independent trials. While the values of $N$ and $P$
were kept constant across these trials, a new set of random patterns
was generated for each trial to ensure that the observed phenomena are
not specific to a single pattern realization.  For the recall
experiments, noisy initial states were generated by randomly flipping
a specified fraction of bits in the target pattern.

\subsection{Empirical Observations: A Fundamental Dichotomy}
\label{sec:sparsity}
To examine the learned representation, we apply both compression
schemes to networks trained on the Ridge. While the main analysis uses
$P/N=3.0$, similar behavior is observed for other settings (e.g.,
$P/N=2.0$; see Appendix~\ref{app:generality}).

\begin{figure}[t]
\begin{center}
  \includegraphics[width=\columnwidth]{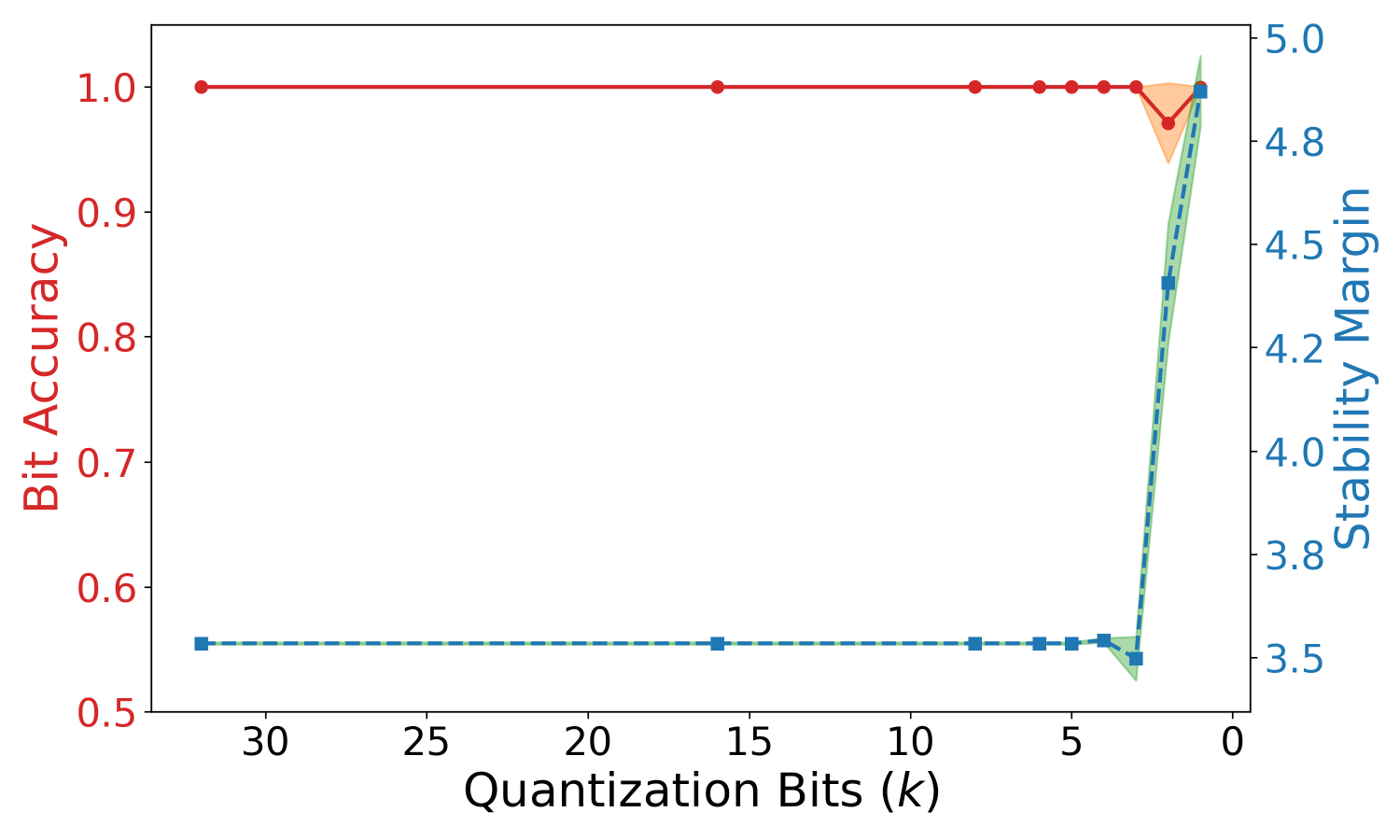}
  \caption{\textbf{Robustness to Uniform Quantization.}  Bit accuracy
    (red solid line) and stability margin (blue dashed line) as a
    function of quantization bit depth $k$.  Shaded areas indicate
    standard deviation over 10 trials.  }
\label{fig:quantization}
\end{center}
\end{figure}

\textbf{Robustness to Quantization:} Figure~\ref{fig:quantization}
shows bit accuracy and stability margin as functions of the
quantization bit depth $k$. The network maintains perfect recall
accuracy even at 2-bit resolution, with negligible variance across
trials. At $k=1$, accuracy remains high.

The stability margin increases at low bit depths. This effect, termed
\textit{Quantization Saturation}, can be interpreted as
saturation-induced amplification: discretization pushes small weights
toward extreme values, increasing the magnitude of the input potential
while preserving its sign.

\begin{figure}[t]
\begin{center}
  \includegraphics[width=\columnwidth]{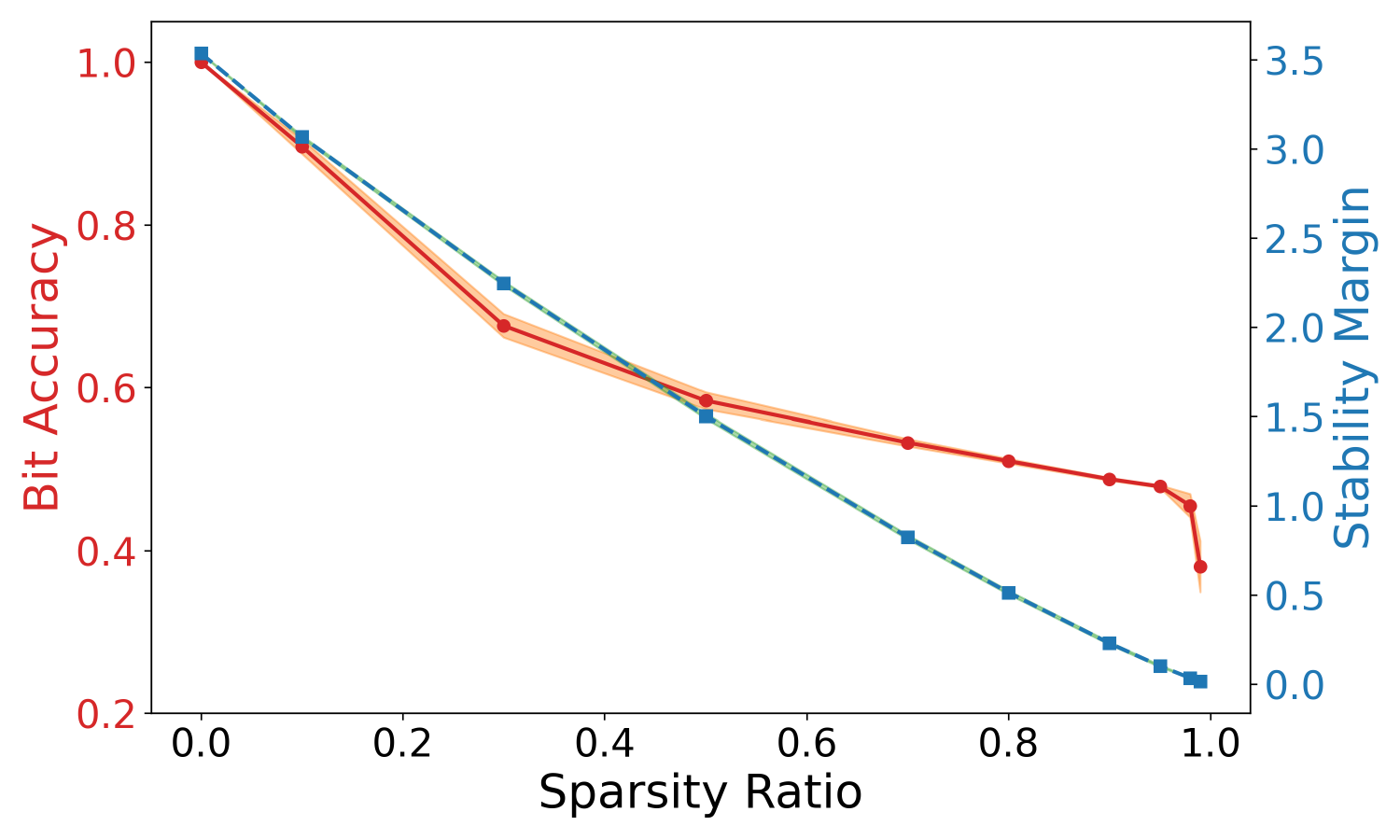}
  \caption{\textbf{Sensitivity to Magnitude Pruning.}  Performance
    metrics as a function of sparsity ratio (fraction of zeroed
    weights).  Shaded areas indicate standard deviation over 10
    trials.  }
\label{fig:sparsity}
\end{center}
\end{figure}

\textbf{Sensitivity to Sparsity:} In contrast,
Figure~\ref{fig:sparsity} shows that performance degrades rapidly as
the sparsity ratio increases. Even moderate pruning (e.g., 10\%) leads
to a significant drop in accuracy, indicating that small-magnitude
weights contribute to the representation.

Taken together, these results indicate qualitatively different
responses to the two compression schemes. The network remains robust
to aggressive quantization, while showing substantial sensitivity to
sparsification. This contrast motivates the theoretical analysis in
the following section.

\subsection{Problem Formulation}
These contrasting responses to the two compression schemes raise a
central question. If the network is robust to quantization, why is it
sensitive to pruning? Addressing this question requires an analysis of
the geometric structure of the learned representation, which is
developed in the next section.

\section{Theoretical Analysis of Representation Geometry}
\label{sec:theory}

In this section, we analyze the geometric structure of the learned
representation from complementary viewpoints. Section~4.1 presents an
interpretation of the emergence of a bimodal weight distribution based
on symmetry breaking. Section~4.2 analyzes the functional structure in
the input space using Walsh influence analysis, clarifying the
distinction between functional sparsity and dense parameterization.
Section~4.3 summarizes the resulting dual representation. Finally,
Section~4.4 provides quantitative support for this interpretation by
examining the scaling behavior of performance degradation under
quantization.

\subsection{Origin of Bimodal Distribution: A Force-Balance Model}
\label{sec:bimodal_origin}

The emergence of a bimodal weight distribution
(Fig.~\ref{fig:weight_dist}), which contributes to robustness against
quantization, can be interpreted through a force-balance model
analogous to symmetry breaking in parameter space. The KLR learning
objective consists of two competing components acting on the dual
variables $\bm{A} = (\alpha_{\mu i})$:

\begin{enumerate}
\item \textbf{Attractive Effect (from $L_2$ Regularization):} The
  weight decay term ($\lambda \|\bm{A}\|_{K}^{2}$) creates a tendency
  that pulls the weights toward the origin ($\bm{A} = \bm{0}$),
  favoring a unimodal Gaussian-like distribution. This corresponds to
  a symmetric state in which the model has no preference for any
  particular feature direction.

\item \textbf{Repulsive Effect (from Margin Maximization):} The
  logistic loss term seeks to maximize the classification margin by
  driving the weights toward large positive or negative values. This
  effect tends to break the symmetry around the origin, separating the
  parameters into positive and negative clusters to ensure pattern
  separability under high load.
\end{enumerate}

During learning on the Ridge of Optimization, we hypothesize that
these two competing effects reach a delicate balance. The repulsive
contribution from the data term becomes sufficiently strong to
overcome the attraction toward zero, potentially rendering the
symmetric state at $\bm{A} = \bm{0}$ unstable. Consequently, the
system is expected to settle into one of two stable non-zero states,
forming an effective double-well-like structure in the weight
distribution.

This interpretation provides a possible explanation for the emergence
of a bimodal distribution without explicit binarization constraints,
in which the parameters effectively take near-discrete values. The
empirical validation of this bimodal distribution is presented in
Sec.~\ref{sec:bimodal_validation}.

\subsection{Holographic vs. Sparse Influence: A Walsh Analysis}
\label{sec:walsh}
\begin{figure*}[t]
  \begin{center}
    \includegraphics[width=\hsize]{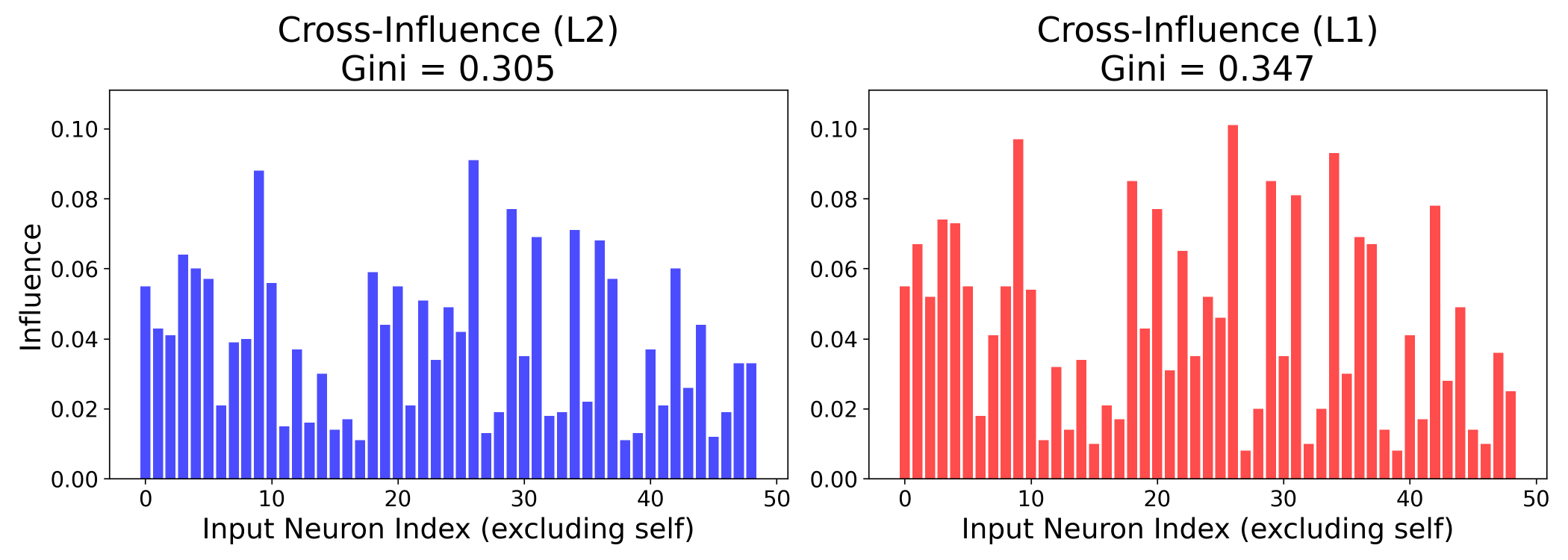}
    \caption{\textbf{Input Influence Distribution for $L_2$ vs. $L_1$
        Regularization}.  The plots show the influence of each input
      neuron on the output of a single target neuron, estimated via
      Walsh analysis.  (Left) The standard KLR model ($L_2$) retains a
      dense tail of cross-influences.  (Right) The $L_1$-regularized
      (Lasso) model shows a much sparser influence profile.  }
    \label{fig:walsh_analysis}
  \end{center}
\end{figure*}
Our results indicate a characteristic property of the memory representation
on the Ridge, which was previously associated with the state of
\textit{Spectral Concentration}~\cite{tamamori2025c}.

The sensitivity to sparsity pruning (Fig.~\ref{fig:sparsity})
motivates a closer examination of the information structure of the
learned representation. To investigate this quantitatively, we employ
Walsh analysis, a set of Fourier-analytic tools for analyzing Boolean
functions~\cite{ODonnell2014}. We treat the update function
$h_{j}(\bm{s})$ for a single output neuron $j$ as a mapping
$\{-1, 1\}^{N} \rightarrow \mathbb{R}$ and compute the influence of
each input neuron $s_{i}$. The influence $I_{i}(h_{j})$ measures the
probability that flipping input $s_{i}$ changes the sign of the output
$h_{j}$, thereby quantifying its functional importance.

We compared the influence distribution of the standard KLR model
($L_2$ regularization) with that of a model trained using
$L_1$ regularization (Lasso), which explicitly promotes sparsity. To
isolate the effect of inter-neuron couplings, we analyzed the
``cross-influence'' by excluding the dominant self-influence term
($i=j$). The results are shown in
Fig.~\ref{fig:walsh_analysis}.

Contrary to a naive holographic interpretation, which would predict a
uniform influence distribution, both models exhibit a sparse influence
structure, indicating that the KLR network performs a form of feature
selection. However, an important quantitative difference appears in
the tail behavior of the influence distribution. To measure the
inequality of the influence distribution, we calculate the Gini
coefficient $G \in [0, 1]$~\cite{Hurley2009}, where $G=0$ indicates a
perfectly uniform distribution and a value approaching $1$ indicates
extreme sparsity. The $L_2$-regularized model (left panel of
Fig.~\ref{fig:walsh_analysis}) retains a richer and denser tail of
small cross-influences, yielding a Gini coefficient of
$G \approx 0.305$. In contrast, the $L_1$-regularized model (right
panel of Fig.~\ref{fig:walsh_analysis}) exhibits a more unequal
distribution with a higher Gini coefficient of $G \approx 0.347$.

These results suggest that, although the effective input-output
mapping exhibits sparse characteristics, the underlying
parameterization of the $L_2$ model remains more distributed.
Together with its robustness to quantization, this behavior is
consistent with the notion of a holographic representation discussed
in Sec.~\ref{sec:related_work}, and helps explain the observed
sensitivity to pruning.

\subsection{Geometric Interpretation: Duality of Representation}
This subsection integrates the preceding analyses into a unified
geometric interpretation of the learned representation. Combining the
analyses of the weight distribution and input influence, our results
suggest a dual structure in the KLR memory representation:

\begin{itemize}
\item \textbf{Parameter Space (Digital \& Dense):} The weights
  $\bm{A}$ self-organize into a bimodal, digital-like structure. This
  representation appears to be dense rather than sparse, which may
  underlie its sensitivity to pruning. At the same time, its intrinsic
  discreteness makes it highly robust to quantization.

\item \textbf{Input Space (Sparse Influence):} The function computed
  by the network exhibits sparse input dependencies. This suggests
  that the network selectively relies on a subset of informative
  features, consistent with effective feature learning.
\end{itemize}

This dual structure provides an interpretation of the contrasting
effects of compression. While the effective input-output mapping
exhibits sparse characteristics, the underlying parameterization
remains distributed. In the sense discussed in
Sec.~\ref{sec:related_work}, this behavior is consistent with a
holographic representation at the parameter level.

The network appears to learn a sparse functional structure in the
input space while implementing it through a dense and distributed
parameterization in the feature space. This separation between
functional sparsity and dense parameterization explains why
quantization preserves performance whereas pruning causes substantial
degradation: quantization largely preserves the coarse distributed
structure of the parameters, while pruning removes components that
contribute to the distributed representation.

More broadly, this separation between sparse functional structure and
dense parameterization may provide a useful perspective for
understanding robust computation in neural systems.

\subsection{Scaling Law of Quantization Error}
\label{sec:scaling_law}

This subsection does not introduce a new theoretical mechanism.
Instead, it provides quantitative support for the geometric
interpretation developed in the preceding subsections by examining how
performance degradation scales with quantization strength.

Theoretically, under a local quadratic approximation of the loss
landscape, the degradation is expected to scale quadratically with the
quantization step size $\Delta$ (see
Appendix~\ref{app:scaling_derivation} for a heuristic derivation). To
test this hypothesis, we measured the degradation of the Stability
Margin for a range of quantization bit depths
$k \in \{16, 12, 10, 8, 7, 6, 5, 4, 3, 2\}$, both on and off the
Ridge. While we focus on the Stability Margin as a sensitive probe,
similar trends were observed for other performance metrics (not
shown).

\begin{figure}[t]
    \centering
    \includegraphics[width=\columnwidth]{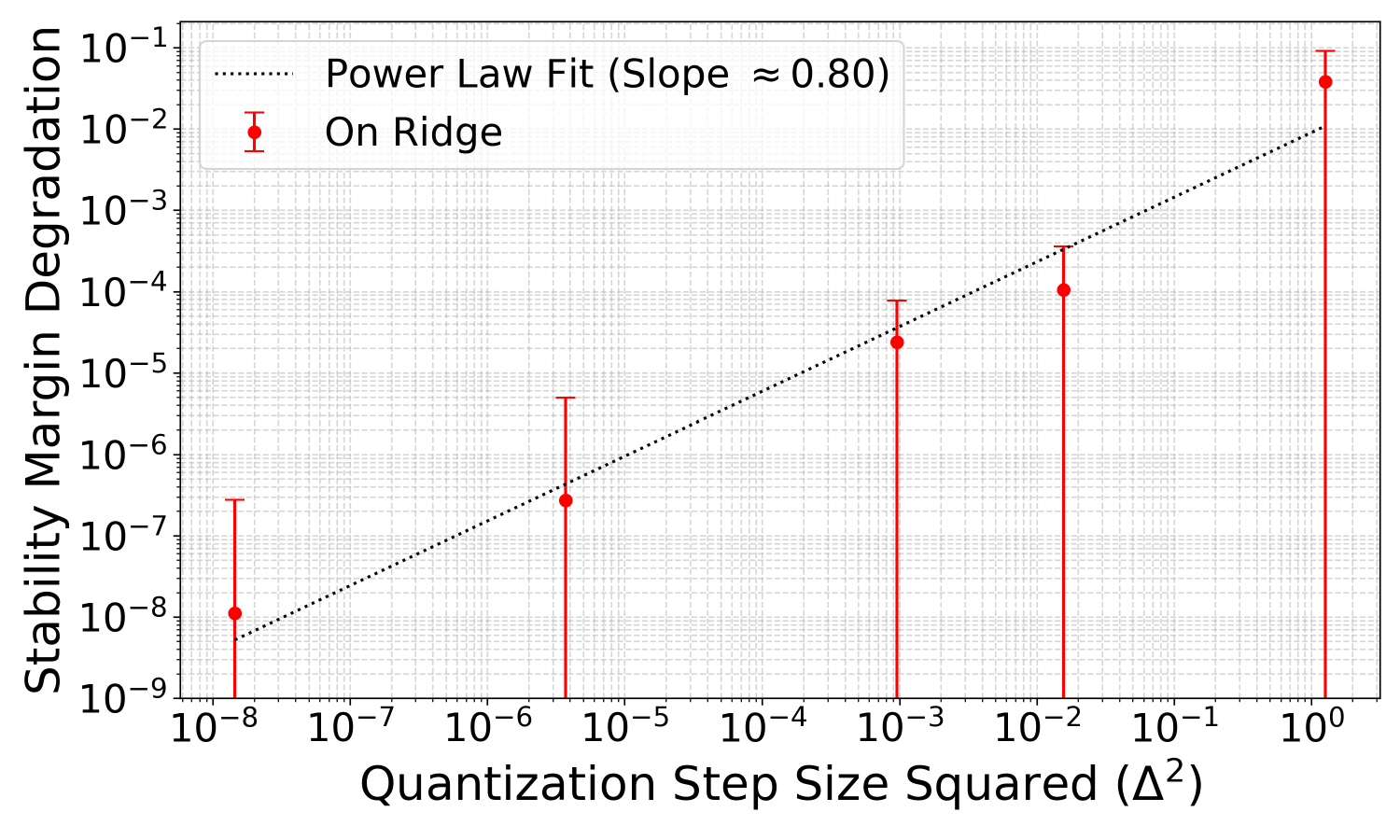}
    \caption{\textbf{Scaling Law of Quantization Error on the Ridge.}
      The plot shows the degradation in Stability Margin as a function
      of the squared quantization step size ($\Delta^2$) on a log-log
      scale. Error bars indicate standard deviation over 10 trials.  }
    \label{fig:scaling_law}
\end{figure}

The results for the ``On Ridge'' regime are shown in
Fig.~\ref{fig:scaling_law}. Data points are visible only where the
degradation is positive; for higher bit depths, the degradation was
negligible (zero or negative) and therefore omitted on the logarithmic
scale. The visible points are consistent with a power-law
relationship. A linear fit to the logarithmic data yields a slope of
approximately $0.80$, giving the empirical scaling relation:
\begin{equation}
    \text{Degradation} \propto (\Delta^2)^\beta, \quad \text{with } \beta \approx 0.80.
\end{equation}
Although this exponent deviates slightly from the ideal quadratic
prediction ($\beta = 1$), the observed power-law scaling indicates
that the degradation progresses in a predictable and gradual manner.
In contrast, for the ``Off Ridge (Local)'' regime, the degradation was
negligible across all bit depths, consistent with the flatter energy
landscape in this region (see Appendix~\ref{app:local_scaling} for the
detailed plot). This scaling analysis supports the interpretation that
the geometric structure of the Ridge governs its robustness to
parameter perturbations.

We conjecture that the deviation from the ideal exponent arises from
finite-size effects and higher-order nonlinearities in the attractor
geometry that are not captured by the quadratic approximation.
Furthermore, the highly skewed spectrum of the Fisher information
matrix reported for deep networks~\cite{Karakida2020} suggests that
the parameter space is strongly anisotropic. In such a geometry,
quantization noise may be preferentially absorbed along flat
directions associated with small eigenvalues, resulting in weaker
degradation than predicted by the quadratic approximation. A more
precise theoretical characterization of this exponent is left for
future work.

\section{Experimental Validation}
\label{sec:validation}
The theoretical analysis in the previous section proposed a dual
structure of the KLR representation: a bimodal structure in parameter
space and sparse influence in the input space. In this section, we
provide further experimental evidence supporting these theoretical
interpretations.
\subsection{Preservation of Basin Depth under Quantization}
Our theory suggests that the bimodal, discrete-like nature of the
weights should make the attractor landscape robust to quantization. To
examine this, we evaluated the noise robustness of the quantized
network.

\begin{figure}[t]
\begin{center}
  \includegraphics[width=\columnwidth]{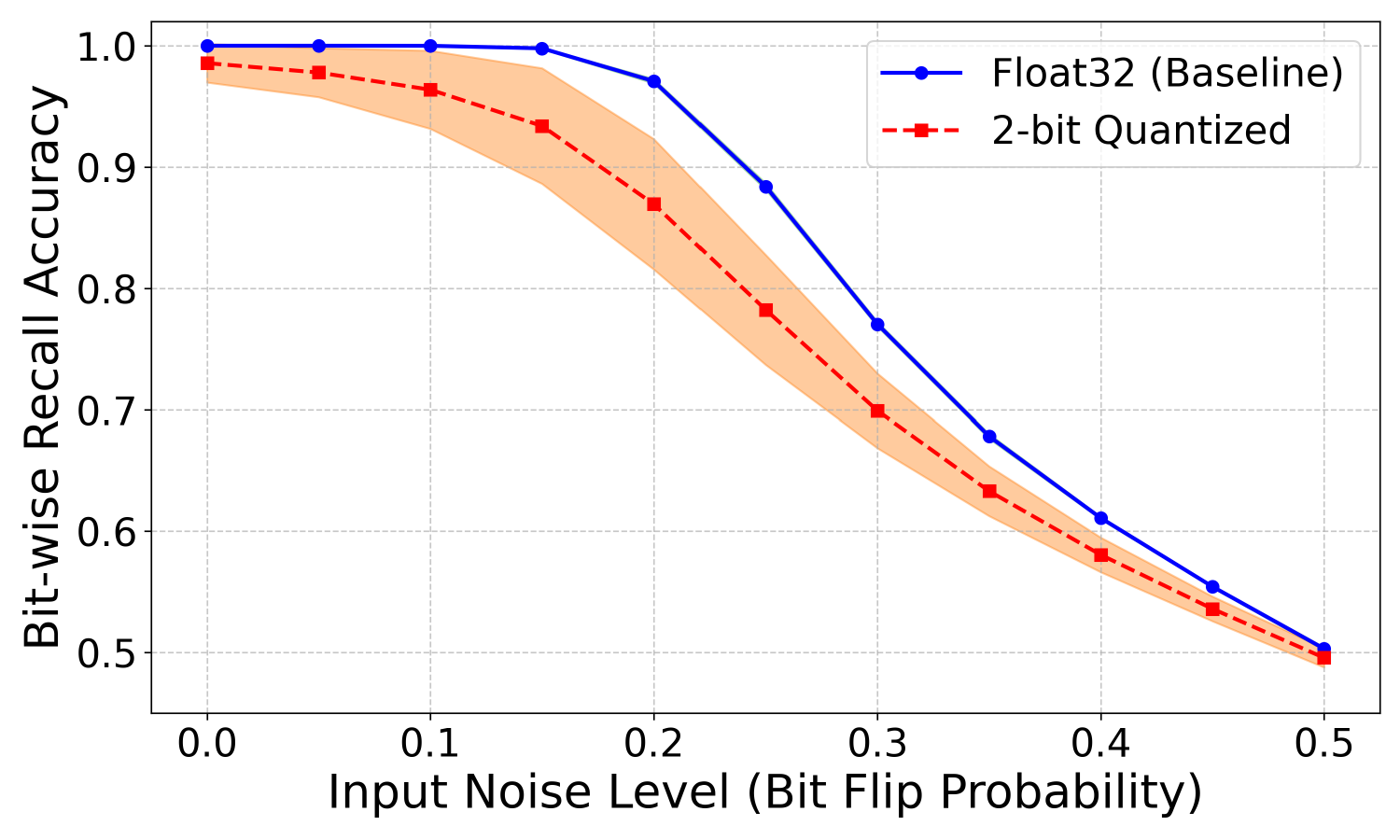}
  \caption{\textbf{Noise Robustness Comparison (Bit-wise Accuracy).}
    Recall performance as a function of input noise level (bit flip
    probability).  Shaded areas indicate standard deviation over 10
    trials.  }
\label{fig:robustness}
\end{center}
\end{figure}

Figure~\ref{fig:robustness} compares the bit-wise recall accuracy of
the full-precision (Float32) and 2-bit quantized models as a function
of input noise level. The performance of the 2-bit model closely
tracks that of the full-precision baseline, maintaining high accuracy
even under significant corruption (e.g., $>85\%$ accuracy at 20\%
noise). These results indicate that quantization largely preserves the
attractor basin structure, consistent with the discrete parameter
organization discussed in Sec.~\ref{sec:bimodal_origin}.

\subsection{Empirical Evidence of Bimodal Weight Distribution}
\label{sec:bimodal_validation}

A central component of our theoretical interpretation is the
spontaneous emergence of a bimodal weight distribution, modeled in
Sec.~\ref{sec:bimodal_origin} as the result of competing effects in
the learning objective. To examine this hypothesis, we analyzed the
empirical distribution of the learned weights $\bm{A}$ on the Ridge
($\gamma=0.02$, $P/N=3.0$).

\begin{figure}[t]
\begin{center}
  \includegraphics[width=\columnwidth]{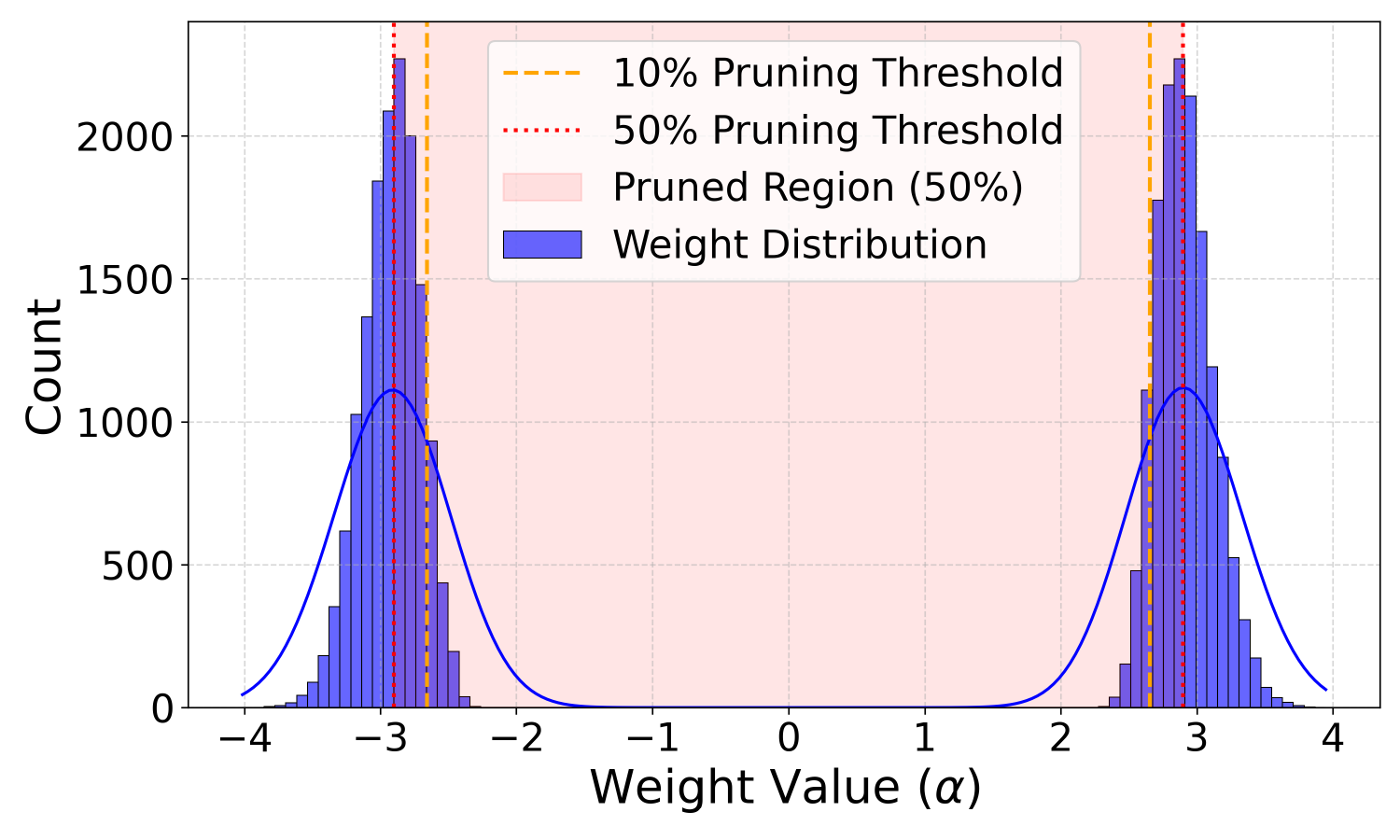}
  \caption{\textbf{Weight Distribution on the Ridge.}  The learned
    weights exhibit a distinct bimodal distribution, concentrating
    around two non-zero values ($\pm 3.0$) with very few weights near
    zero.  }
  \label{fig:weight_dist}
\end{center}
\end{figure}

Figure~\ref{fig:weight_dist} shows the histogram of the weight
elements. Consistent with the proposed force-balance interpretation,
the distribution is not unimodal. Instead, it separates into two
clusters centered around non-zero values (approximately $\pm 3.0$),
with relatively few weights near the origin. This observation is
consistent with the interpretation of an effective double-well-like
structure governing the weight distribution.

Furthermore, the observed distribution provides an intuitive
explanation for the sensitivity to magnitude pruning. Since only a
small fraction of weights are located near zero, pruning
preferentially removes weights from the tails of the two clusters,
where they still contribute to the distributed representation.

\subsection{The Ridge as a Robust Operating Region}
\label{sec:distance_to_ridge}
Our theoretical interpretation suggests that the geometric structure
of the Ridge plays a central role in quantization robustness. To
examine this hypothesis, we investigated how quantization sensitivity
changes as the kernel parameter moves away from the Ridge center
($\gamma^{*} = 0.02$). Specifically, we measured the accuracy
degradation induced by 2-bit quantization for various $\gamma$ values
while fixing the load at $P/N = 3.0$.

\begin{figure}[t]
\centering
\includegraphics[width=\columnwidth]{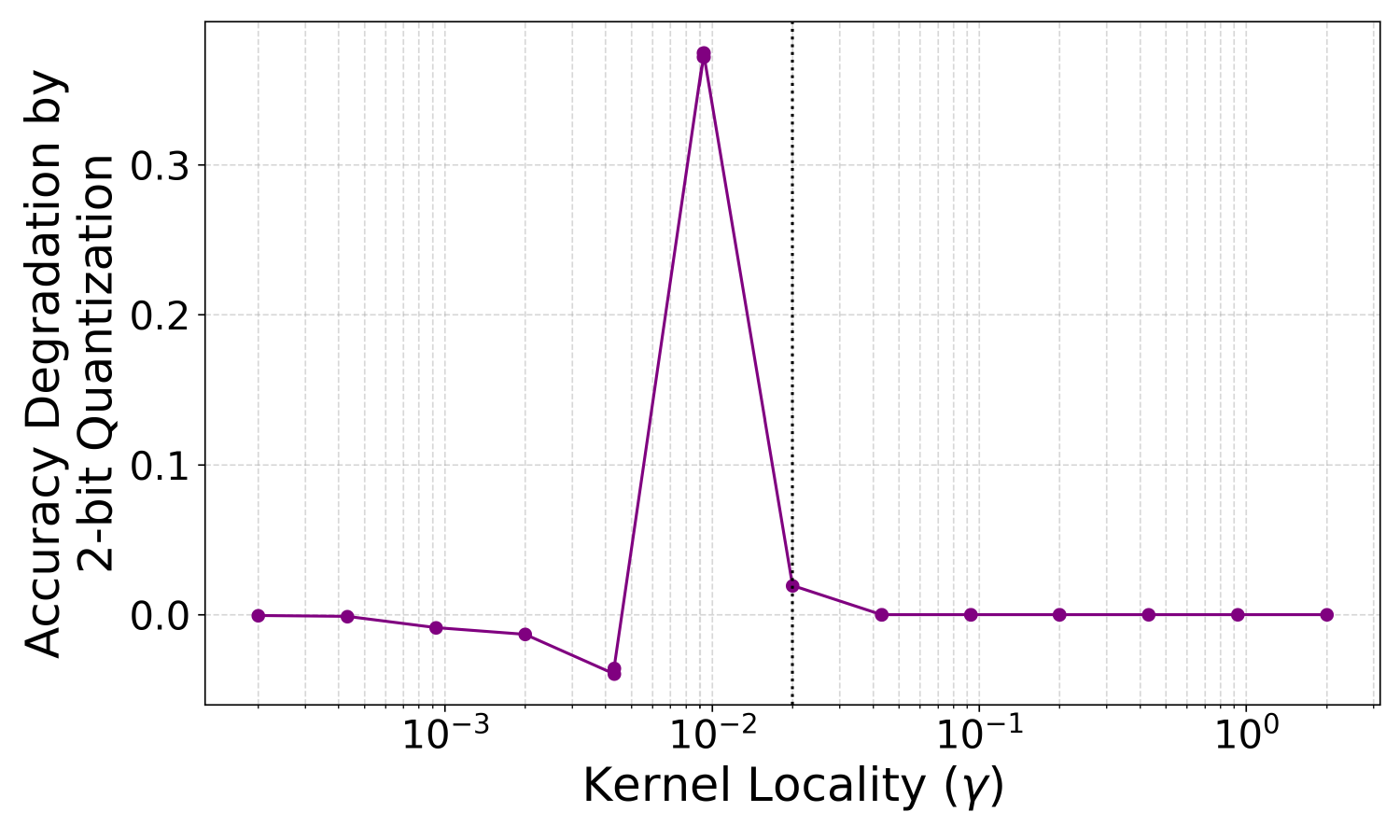}
\caption{\textbf{Quantization Sensitivity vs. Kernel Locality.}  The
  plot shows the accuracy degradation caused by 2-bit quantization as
  a function of the kernel locality parameter $\gamma$.  The Ridge
  center ($\gamma^{*} = 0.02$, vertical dotted line) is located in a
  robust region.  }
\label{fig:distance_to_ridge}
\end{figure}

The results are shown in Fig.~\ref{fig:distance_to_ridge}. The plot
reveals an asymmetric valley-like structure. In the local regime
($\gamma > \gamma^{*}$), the accuracy degradation remains nearly zero,
consistent with the smaller weight magnitudes in this regime and their
reduced sensitivity to discretization. In contrast, the most
pronounced feature appears in the global regime
($\gamma < \gamma^{*}$). As $\gamma$ decreases from the Ridge center,
the degradation rapidly increases and reaches a sharp peak around
$\gamma \approx 0.01$, corresponding to catastrophic retrieval
failure.

This peak likely corresponds to the capacity boundary of the network.
Near this critical region, the system becomes highly sensitive to
parameter perturbations, and quantization can induce a complete
collapse of retrieval performance. The region further to the left
($\gamma < 0.01$) corresponds to an overloaded phase in which even the
full-precision model fails to learn stable memories properly.

Overall, these results suggest that the Ridge of Optimization is not
merely a point of minimal quantization error, but rather a robust
operating region located near the transition to memory overload.

\section{Discussion}
\label{sec:discussion}
Our analysis, combining empirical observations with theoretical
modeling and experimental validation, suggests that KLR networks
employ a structured form of information representation. The present
results were obtained near the Ridge of Optimization, which we regard
as a structurally important regime for high-capacity kernel models.
To contextualize these findings, it is helpful to distinguish which
observed behaviors appear specific to the Ridge and which may reflect
more general properties of KLR learning.

Based on our theoretical framework, the emergence of the bimodal
weight distribution and the resulting robustness to quantization are
phenomena specific to the Ridge. As discussed in
Sec.~\ref{sec:bimodal_origin}, these features rely on a delicate
balance between margin maximization and regularization forces, a
balance that is disrupted outside this optimal regime.  Conversely, we
hypothesize that the underlying dense parameterization, which leads to
sensitivity to pruning, is a more general characteristic of KLR-based
associative memories. Because the logistic loss function assigns
non-zero weights to all training patterns (unlike the strict sparsity
in support vector machines), the memory representation tends to remain
distributed across the parameter space. However, a comprehensive
empirical verification of the compression behavior across all
optimization regimes remains an open question for future study.

\subsection{The Duality of Representation}
The central theoretical finding of this paper is the dual nature of
the memory representation. The Walsh analysis (Sec.~\ref{sec:walsh})
showed that the network learns a sparse input mapping, effectively
performing feature selection. In contrast, the parameter
representation in the feature space remains dense and holographic, as
indicated by its sensitivity to pruning
(Sec.~\ref{sec:sparsity}). This duality suggests a trade-off between
efficiency and robustness: the network identifies a relatively small
set of informative features while encoding them in a distributed and
fault-tolerant manner. This principle may extend beyond associative
memory to more general forms of neural computation.

\subsection{Spectral Bias and Memory Stability}
Recent studies by Karakida et al.~\cite{Karakida2020, Karakida2021}
have identified a structural property commonly observed in deep neural
networks: the Fisher information matrix often exhibits a highly skewed
spectrum dominated by a small number of large eigenvalues.  Although
such spectral structure is sometimes described as ``pathological,''
our results suggest an alternative interpretation in the context of
associative memory.

We hypothesize that KLR learning near the Ridge accentuates this
intrinsic spectral bias. By exploiting the dominant spectral modes,
the system may efficiently form deep and stable attractor basins,
while suppression of the bulk spectrum reduces inter-pattern
interference. From this perspective, a spectral bias that may be
detrimental in other learning contexts instead becomes advantageous
for associative memory storage. This property may also reflect a
physiological principle underlying robust and high-capacity memory.
This interpretation further motivates the biological perspective
discussed in Sec.~\ref{sec:bio_plausibility}.

\subsection{The Ridge as a Critical Operating Regime}
Our scaling analysis (Sec.~\ref{sec:scaling_law}) and
distance-to-ridge analysis (Sec.~\ref{sec:distance_to_ridge}) provide
a more nuanced characterization of the Ridge of Optimization. The
Ridge is not merely a point of maximal stability, but rather a
critical operating regime that balances multiple trade-offs. The
observed power-law scaling suggests that the landscape is
geometrically smooth, allowing gradual performance degradation under
quantization. At the same time, the proximity of the Ridge to the
capacity boundary suggests that the system self-organizes near a
transition between stable retrieval and memory failure, a behavior
reminiscent of Self-Organized Criticality~\cite{Beggs2003}.

From this perspective, the Ridge may represent a favorable operating
region that combines high performance with robustness against
catastrophic degradation. Practically, this state may be identified by
monitoring geometric metrics such as Pinnacle
Sharpness~\cite{tamamori2025c}, or more simply, by observing the
emergence of a bimodal weight distribution during training.

\subsection{Revisiting Biological Plausibility}
\label{sec:bio_plausibility}
The holographic and low-precision characteristics of the KLR
representation are consistent with several principles of neural coding
in biological systems. Neural systems are thought to operate robustly
despite noisy and low-precision synaptic components~\cite{Rieke1999};
for example, hippocampal synapses have been estimated to possess a
precision of only a few bits~\cite{Bartol2015}. Our results suggest
that the combination of a dense and distributed representation
(holographic) with a low-precision digital format (e.g., 2-bit
quantization) provides a robust mechanism for memory representation.

Although the KLR learning algorithm itself is not biologically
plausible, the resulting geometric structure, namely a bimodal and
fault-tolerant weight distribution, may reflect a convergent solution
shared by both artificial and biological learning systems.

\subsection{Practical Implications for Hardware}
The findings of this study suggest several implications for
hardware-efficient implementation. In particular, the combination of
quantization robustness and a dense holographic representation enables
substantial reductions in memory and computational cost. Robustness to
2-bit quantization yields a 16-fold reduction in memory footprint
relative to standard 32-bit floating-point storage. For example, a
network with $N = 1000$ and $P = 3000$ would normally require
approximately 12\,MB of memory, whereas 2-bit quantization reduces
this requirement to approximately 0.75\,MB. This reduction may allow
the entire model to fit within the on-chip SRAM of low-power edge
devices.

Furthermore, the retrieval computation can be accelerated through
low-precision arithmetic. By quantizing weights to 1-bit or 2-bit,
floating-point multiply-accumulate operations can be replaced by
bitwise XNOR and population count operations. Such binary operations
have been reported to achieve substantial computational
efficiency~\cite{Rastegari2016}. While sparsity is often used to
reduce computational cost, our results suggest that, for high-capacity
associative memory, low-precision dense computation may provide a more
favorable trade-off between computational efficiency and memory
performance.

In addition to reducing memory and computational cost, quantization
may also improve energy efficiency. A 32-bit floating-point
multiplication consumes approximately 3.7\,pJ of energy on a 45\,nm
process, whereas a binary XNOR operation consumes less than
0.1\,pJ~\cite{Horowitz2014}. For large-scale associative memory with
millions of parameters, this difference may translate into substantial
reductions in power consumption, which is particularly important for
battery-operated edge devices.

\subsection{Limitations and Future Work}
The main text of this study focuses on the recall of random binary
patterns. However, we have verified that the fundamental ``duality''
principle proposed here holds even when storing continuous, non-binary
patterns (see Appendix~\ref{app:non_binary}), confirming that the
quantization robustness is not simply a consequence of binary coding.
How the structural correlations in real-world data (e.g., images or
language representations) may affect the ``duality'' principle
proposed here remains an important subject for future investigation.
We hypothesize that for highly correlated data such as images, the
network would rely even more heavily on a sparse subset of critical
features (e.g., low-frequency components or structural edges) to
distinguish patterns, thereby exacerbating the sparse functional
mapping observed in our Walsh analysis. Conversely, the dense,
holographic nature of the underlying parameter space is expected to be
preserved, maintaining robustness against quantization. Furthermore,
the present analysis is limited to post-training
quantization. Extending the proposed geometric framework to
quantization-aware training is another important direction for future
work.

\section{Conclusion}
\label{sec:conclusion}
In this work, we presented a geometric interpretation of the
compression robustness observed in high-capacity KLR Hopfield
networks. Our analysis showed that the contrasting responses to
quantization and sparsity arise from a dual representation: a dense,
bimodal structure in parameter space and a sparse functional mapping
in the input space.

From a theoretical perspective, we interpreted the emergence of this
structure through a spontaneous symmetry breaking mechanism.
Empirically, quantitative scaling analysis revealed a predictable
power-law relationship governing the degradation induced by
quantization. Our findings further suggest that the ``Ridge of
Optimization'' is not merely a high-performance region, but a critical
operating regime located near the transition to memory overload. In
this regime, the network combines geometric stability with high memory
capacity. The principle of ``sparse function, dense representation''
may therefore provide a useful perspective for understanding robust
and efficient computation in both artificial and biological neural
systems.

Finally, the present results suggest that kernel-based associative
memories may be implemented efficiently on low-power hardware through
low-precision quantization. Significant reductions in memory usage and
computational cost are achievable while largely preserving retrieval
performance. These findings indicate that KLR Hopfield networks are a
promising candidate for neuromorphic and edge-oriented memory systems.

\funding
Not applicable.

\conflictsofinterest
The author declares no competing interests.

\authorcontribution
The sole author contributed to the present work.

\aitools
The author used ChatGPT (GPT-5.3) and Gemini 2.5 Pro for proofreading
the English manuscript.

\appendix
\section{Heuristic Derivation of Quadratic Scaling}
\label{app:scaling_derivation}

Here, we provide a heuristic argument for why the performance
degradation is expected to scale quadratically with the quantization
step size, i.e., $\mathrm{Degradation} \propto \Delta^{2}$.

Let $L(\bm{A})$ denote the loss function. Under the assumption that
the loss landscape is smooth and locally approximated by a second-order
Taylor expansion around an optimum $\bm{A}^{*}$,
\begin{equation}
  L(\bm{A}^{*} + \boldsymbol{\delta})
  \approx
  L(\bm{A}^{*})
  + \nabla L(\bm{A}^{*})^{\top} \boldsymbol{\delta}
  + \frac{1}{2} \boldsymbol{\delta}^{\top} \bm{H} \boldsymbol{\delta},
\end{equation}
where $\boldsymbol{\delta} = Q_{k}(\bm{A}) - \bm{A}^{*}$ denotes the
quantization error vector and $\bm{H}$ is the Hessian matrix (or the
Fisher information matrix). Since $\bm{A}^{*}$ corresponds to a local
minimum, the gradient term $\nabla L(\bm{A}^{*})$ is approximately
zero. The increase in loss is therefore approximated as
\begin{equation}
  \Delta L
  =
  L(Q_{k}(\bm{A})) - L(\bm{A}^{*})
  \approx
  \frac{1}{2} \boldsymbol{\delta}^\top \bm{H} \boldsymbol{\delta}.
\end{equation}

Uniform quantization introduces a rounding error for each parameter,
which is uniformly distributed over the interval
$[-\Delta/2, \Delta/2]$. The mean squared error for a single parameter
is therefore
\begin{equation}
  \mathbb{E}[\delta_i^2]
  =
  \frac{1}{\Delta}
  \int_{-\Delta/2}^{\Delta/2} x^2 \, dx
  =
  \frac{\Delta^2}{12}.
\end{equation}

Assuming that the quantization errors are uncorrelated across the
$D = P \times N$ dimensions, the expected increase in loss becomes
\begin{align}
  \mathbb{E}[\Delta L]
  &\approx
  \frac{1}{2}
  \mathbb{E}
  \left[
    \sum_i \sum_j H_{ij} \delta_i \delta_j
  \right]
  \nonumber\\
  &=
  \frac{1}{2}
  \sum_i H_{ii} \mathbb{E}[\delta_i^2]
  \nonumber\\
  &=
  \frac{1}{2}
  \mathrm{Tr}(\bm{H})
  \frac{\Delta^2}{12}.
\end{align}

If the performance degradation (e.g., the loss of stability margin)
is monotonically related to the increase in loss, then the degradation
is expected to scale proportionally with $\Delta^{2}$:
\begin{equation}
  \mathrm{Degradation}
  \propto
  \mathbb{E}[\Delta L]
  \propto
  \Delta^2.
\end{equation}

Despite its simplifying assumptions, this model provides a heuristic
interpretation of the power-law scaling observed in our experiments
and explains the origin of the quadratic dependence.

\section{Generality Across Storage Loads}
\label{app:generality}

\begin{figure}[t]
\centering
\begin{minipage}{0.5\textwidth}
\centering
\includegraphics[width=\linewidth]{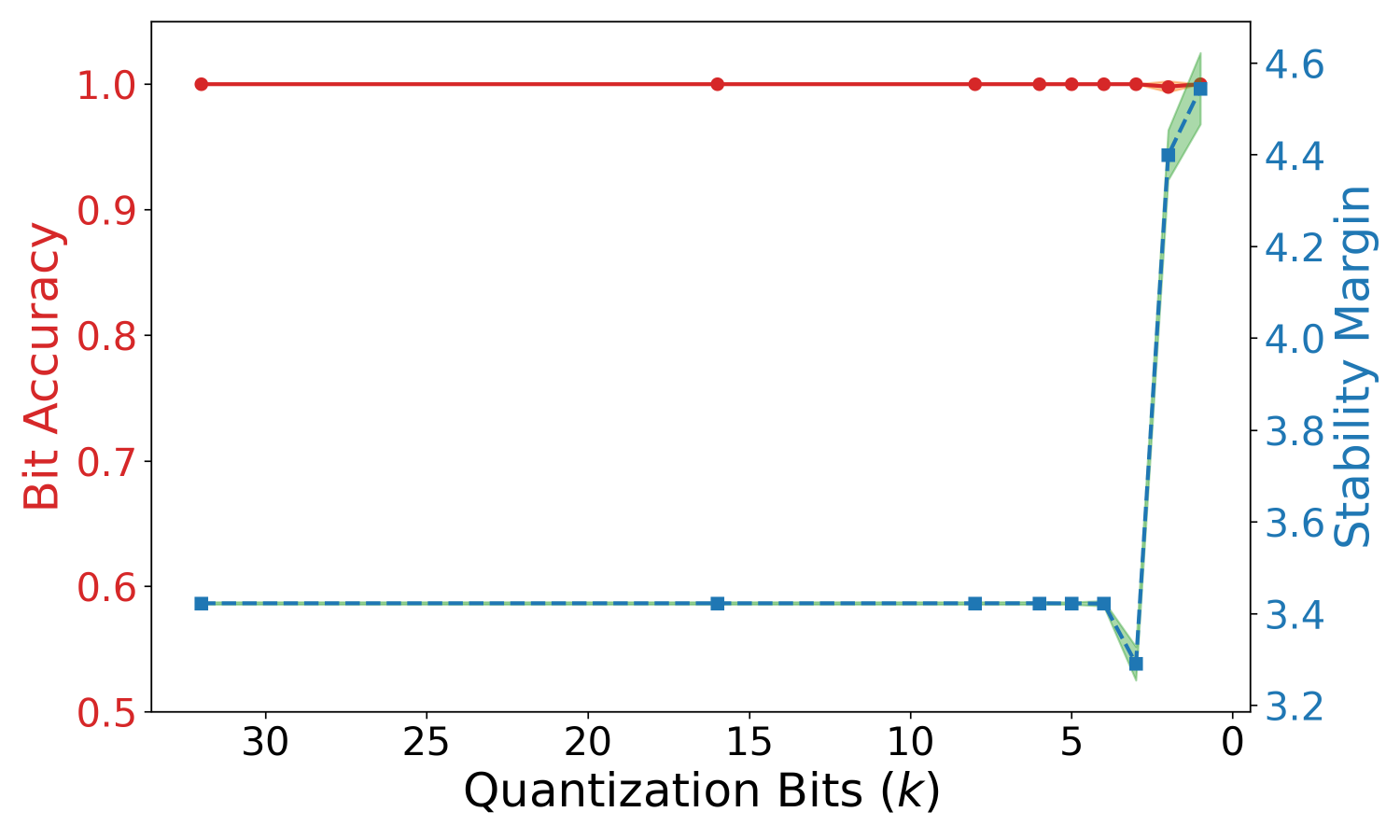}
\centerline{(a) Quantization Robustness}
\vspace{0.5mm}
\end{minipage}
\begin{minipage}{0.5\textwidth}
\centering
\includegraphics[width=\linewidth]{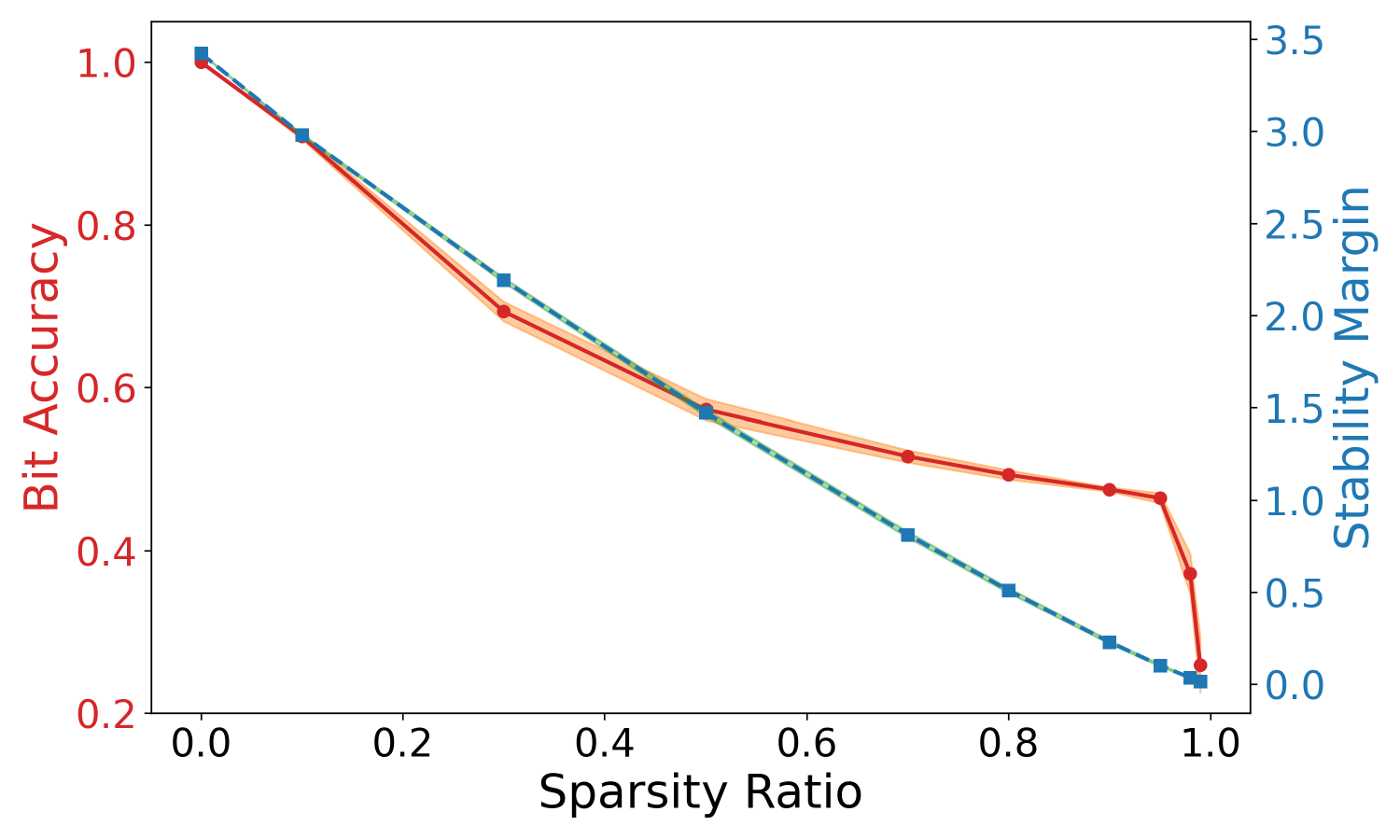}
\centerline{(b) Sparsity Sensitivity}
\end{minipage}
\caption{\textbf{Performance of the compressed network at a lower
    storage load of $P/N=2.0$}.  Shaded areas indicate standard
  deviation over 10 trials.  }
\label{fig:compression_pn2}
\end{figure}

To confirm that the observed behavior is not specific to the high-load
condition of $P/N = 3.0$, we repeated the compression experiments at a
lower, yet still challenging, storage load of $P/N = 2.0$. The
results, shown in Fig.~\ref{fig:compression_pn2}, exhibit the same
qualitative trends.

The network maintains perfect bit accuracy down to 2-bit quantization,
whereas performance degrades rapidly under sparsity pruning. The
weight distribution (not shown) also remains bimodal.  These results
suggest that the ``sparse function, dense representation'' structure
is a robust property of the Ridge of Optimization across multiple
high-load conditions, rather than an artifact of a particular
hyperparameter setting.

\section{Quantization Behavior in the Local Regime}
\label{app:local_scaling}

\begin{figure}[t]
  \centering
  \includegraphics[width=\columnwidth]{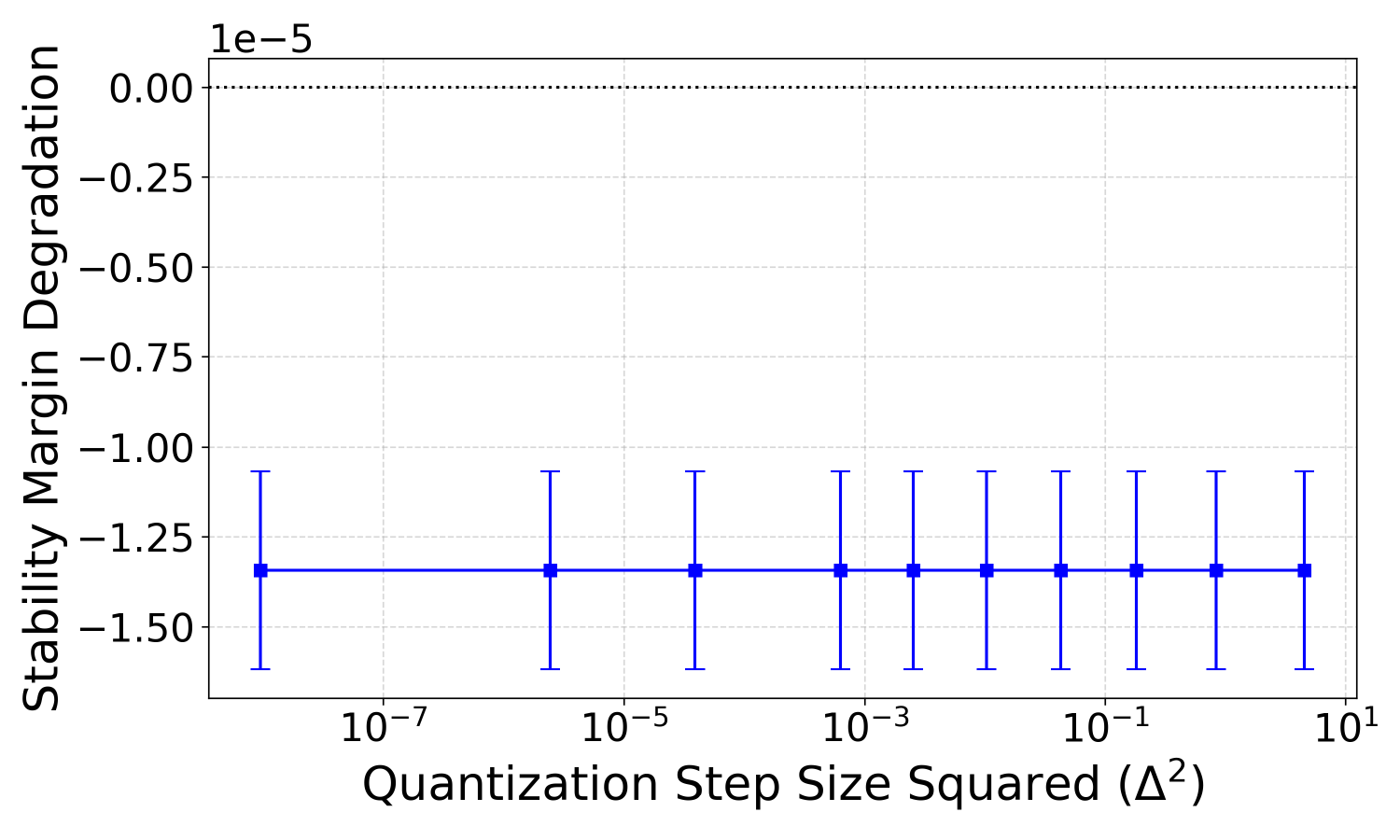}
  \caption{\textbf{Stability margin degradation in the Local regime
      ($\gamma=0.1$)}. The data points show the mean degradation over
    10 trials, and the error bars indicate the standard deviation. The
    degradation remains negligible across all tested quantization
    levels.}
  \label{fig:local_scaling_plot}
\end{figure}

To complement the scaling analysis presented in
Sec.~\ref{sec:scaling_law}, this appendix provides the detailed
results for the ``Off Ridge (Local)'' regime. As shown in
Fig.~\ref{fig:local_scaling_plot}, the degradation of the stability
margin remains close to zero (or slightly negative) across all tested
bit depths. This indicates that the network in the local regime is
largely insensitive to parameter quantization, in contrast to the
predictable power-law degradation observed near the Ridge.

\section{Robustness with Non-Binary Continuous Patterns}
\label{app:non_binary}
\begin{figure}[t]
    \centering
    \begin{minipage}{0.5\textwidth}
        \centering
        \includegraphics[width=\linewidth]{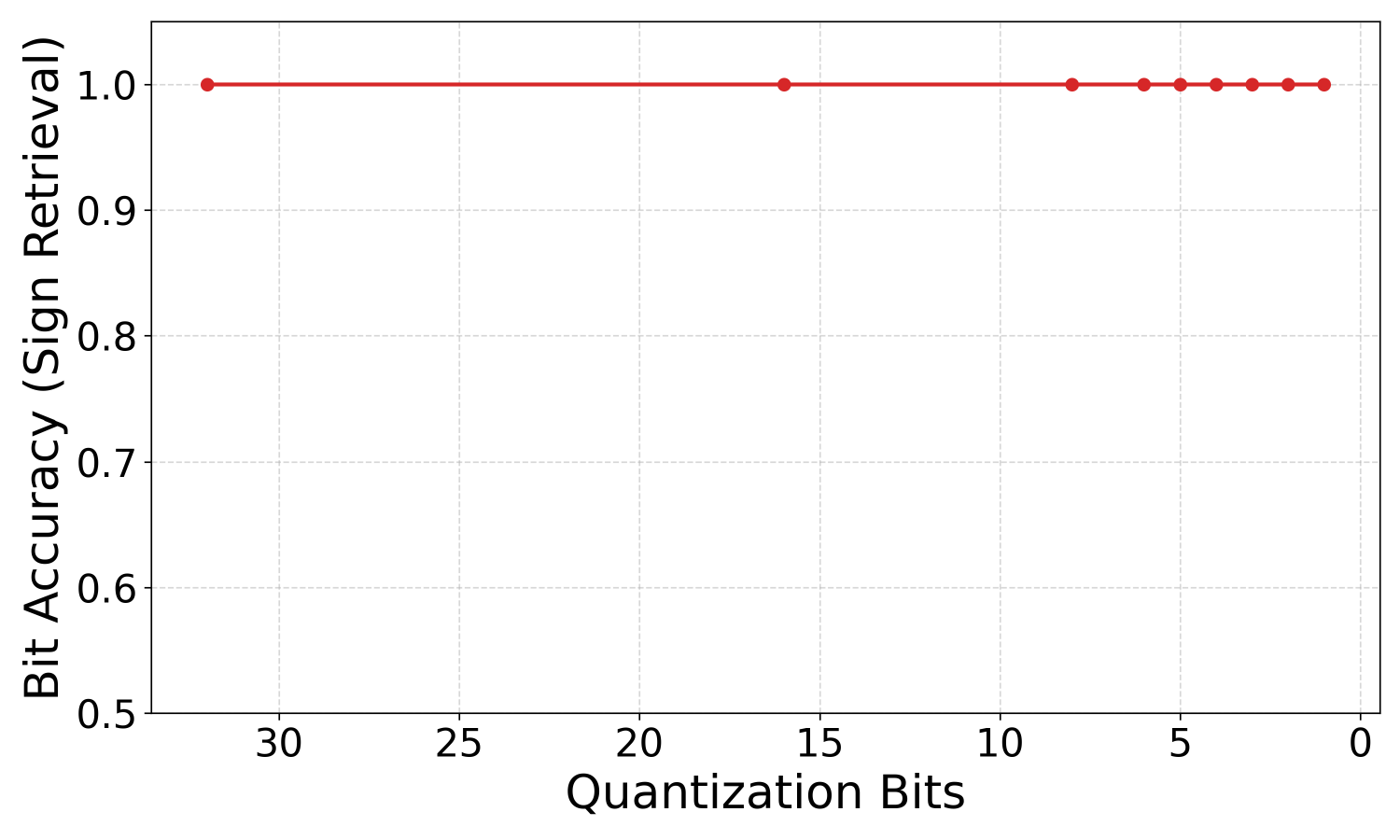}
        \centerline{(a) Quantization Robustness}
        \vspace{0.2mm}
    \end{minipage}
    \hfill
    \begin{minipage}{0.5\textwidth}
        \centering
        \includegraphics[width=\linewidth]{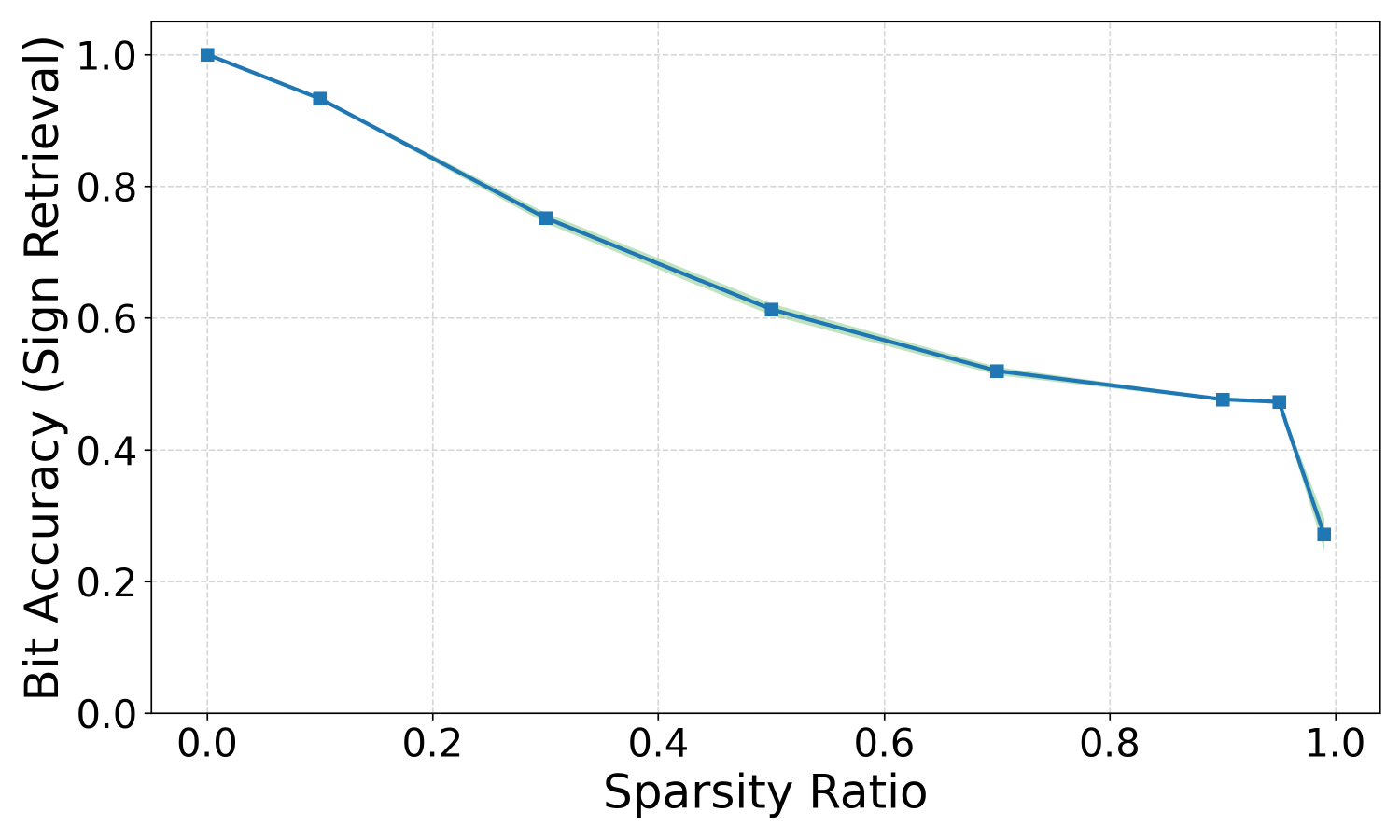}
        \centerline{(b) Sparsity Sensitivity}
      \end{minipage}
      \caption{\textbf{Compression Robustness for Non-Binary
          Continuous Patterns ($P/N=3.0, \gamma=0.1$).}  (a) The
        network achieves perfect sign retrieval even under 1-bit
        quantization. (b) Performance degrades monotonically under
        sparsity pruning. Shaded areas indicate standard deviation
        over 10 trials. Note that the variance is so small that the
        shaded bands are largely obscured by the plotted lines.}
    \label{fig:non_binary_compression}
\end{figure}

To confirm that the observed extreme robustness to quantization is not
merely an artifact of the simplicity of binary coding ($\{-1, 1\}$),
we conducted additional experiments using continuous-valued input
patterns.  Here, each element of the stored patterns
$\boldsymbol{\xi}^\mu$ was sampled independently from a uniform
distribution over $[-1, 1]$. The target states for the KLR learning
were defined by the signs of these continuous patterns, maintaining
the standard Hopfield retrieval objective
$\bm{s} = \mathrm{sign}(\bm{h})$.  Due to the smaller average squared
norm of uniformly distributed continuous vectors
($\mathbb{E}[\|\boldsymbol{\xi}\|^2] \approx N/3$) compared to binary
vectors ($N$), the optimal kernel locality shifts. We empirically
found that $\gamma = 0.1$ places the network on the corresponding
Ridge of Optimization, achieving perfect initial recall accuracy at
$P/N=3.0$.

Figure~\ref{fig:non_binary_compression} presents the performance of
the compressed networks under this continuous pattern
regime. Consistent with our previous findings, the network exhibits
the same qualitative dichotomy observed with binary patterns: it
maintains perfect sign retrieval accuracy even when the dual variables
are quantized down to 1 bit (Fig.~\ref{fig:non_binary_compression}
(a)), while its performance degrades rapidly under magnitude pruning
(Fig.~\ref{fig:non_binary_compression} (b)). This confirms that the
dense, holographic nature of the memory representation is an intrinsic
geometric property of KLR learning on the Ridge, independent of the
discrete or continuous nature of the input data.

\end{document}